\documentclass[runningheads]{llncs}

\usepackage{eccv}

\usepackage{eccvabbrv}

\usepackage{graphicx}
\usepackage{booktabs}
\usepackage{url}

\usepackage[accsupp]{axessibility}  %

\usepackage{hyperref}

\usepackage{orcidlink}

\definecolor{cvprblue}{rgb}{0.21,0.49,0.74}
\usepackage{algorithm}
\usepackage{algpseudocode}
\usepackage{epsfig}
\usepackage{amsmath}
\usepackage{amssymb}
\usepackage{amsfonts}
\usepackage{multirow}
\usepackage{colortbl}
\usepackage[dvipsnames]{xcolor}
\usepackage{arydshln} %
\usepackage{booktabs}
\usepackage{footmisc}
\usepackage{subcaption}
\usepackage{multirow}
\usepackage{nicematrix} %
\usepackage{tablefootnote}
\definecolor{citecolor}{HTML}{0071bc}
\usepackage{wrapfig,lipsum,booktabs}
\usepackage{pifont}
\usepackage{tikz}

\newlength\savewidth\newcommand\shline{\noalign{\global\savewidth\arrayrulewidth
  \global\arrayrulewidth 1pt}\hline\noalign{\global\arrayrulewidth\savewidth}}
\newcommand{\tablestyle}[2]{\setlength{\tabcolsep}{#1}\renewcommand{\arraystretch}{#2}\centering\footnotesize}
\renewcommand{\paragraph}[1]{\vspace{1.25mm}\noindent\textbf{#1}}

\newcolumntype{x}[1]{>{\centering\arraybackslash}p{#1pt}}
\newcolumntype{y}[1]{>{\raggedright\arraybackslash}p{#1pt}}
\newcolumntype{z}[1]{>{\raggedleft\arraybackslash}p{#1pt}}

\newcommand{\app}{\raise.17ex\hbox{$\scriptstyle\sim$}}

\definecolor{deemph}{gray}{0.6}

\definecolor{baselinecolor}{gray}{.9}
\newcommand{\baseline}[1]{\cellcolor{baselinecolor}{#1}}
\definecolor{zeroshotcolor}{gray}{.3}

\definecolor{LightCyan}{rgb}{0.92,1,1}

\def \CCThree          {\textit{CC3M}\xspace}

\definecolor{demphcolor}{RGB}{144,144,144}
\newcommand{\demph}[1]{\textcolor{demphcolor}{#1}}

\definecolor{LightCyan}{rgb}{0.92,1,1}
\definecolor{darkergreen}{RGB}{21, 152, 56}
\definecolor{red2}{RGB}{252, 54, 65}

\definecolor{bluebell}{rgb}{0.84, 0.84, 0.92}
\newcommand*\colorcmark[1]{%
  \expandafter\newcommand\csname #1cmark\endcsname{\textcolor{#1}{\ding{51}}}%
}
\colorcmark{green}

\newcommand*\colorxmark[1]{%
  \expandafter\newcommand\csname #1xmark\endcsname{\textcolor{#1}{\ding{55}}}%
}

\definecolor{Highlight}{HTML}{39b54a}  %

\newcommand{\hl}[1]{\textcolor{Highlight}{#1}}

\colorxmark{red}
\hypersetup{
    colorlinks=true,
    linkcolor=black,
    citecolor=cyan,
    filecolor=magenta,      
    urlcolor=magenta,
    }

\begin{document}

\title{Dataset Growth} 

\titlerunning{Dataset Growth}

\authorrunning{Z Qin, Z Xu, H Yao, K Wang, and Y You et al.}

\author{Ziheng Qin$^\star$\inst{1}\orcidlink{0009-0001-8571-1228} \and Zhaopan Xu\thanks{equal contribution, zihengq@comp.nus.edu.sg, 20b903054@stu.hit.edu.cn.}\inst{1,2} \and
Yukun Zhou\inst{1} \and Zangwei Zheng\inst{1} \and Zebang Cheng\inst{3} \and Hao Tang\inst{4}
\and Lei Shang\inst{5} \and Baigui Sun\inst{5} 
\and Xiaojiang Peng\inst{3} \and Radu Timofte\inst{4} \and Hongxun Yao$^\dagger$\inst{2} \and Kai Wang$^\dagger$\inst{1} \and Yang You\thanks{corresponding authors, h.yao@hit.edu.cn, \{kai.wang, youy\}@comp.nus.edu.sg}\inst{1}\\
\vspace{5pt}
Code: \url{https://github.com/NUS-HPC-AI-Lab/InfoGrowth}}
\institute{
  \makebox[270pt]{National University of Singapore \and Harbin Institute of Technology}
  \and
  \makebox[270pt]{Shenzhen Technology University \and ETH Zurich \and Alibaba Group}
}

\maketitle

\begin{abstract}
Deep learning benefits from the growing abundance of available data.
Meanwhile, efficiently dealing with the growing data scale has become a challenge. Data publicly available are from different sources with various qualities, and it is impractical to do manual cleaning against noise and redundancy given today's data scale. 
There are existing techniques for cleaning/selecting the collected data. However, these methods are mainly proposed for offline settings that target one of the cleanness and redundancy problems. In practice, data are growing exponentially with both problems. This leads to repeated data curation with sub-optimal efficiency.
To tackle this challenge, we propose InfoGrowth, an efficient online algorithm for data cleaning and selection, resulting in a growing dataset that keeps up to date with awareness of cleanliness and diversity. InfoGrowth can improve data quality/efficiency on both single-modal and multi-modal tasks, with an efficient and scalable design. Its framework makes it practical for real-world data engines.
\end{abstract}
    
\section{Introduction}
\label{sec:intro}

\begin{figure*}[t]
\begin{subfigure}{0.48\textwidth}
\includegraphics[width=\linewidth]{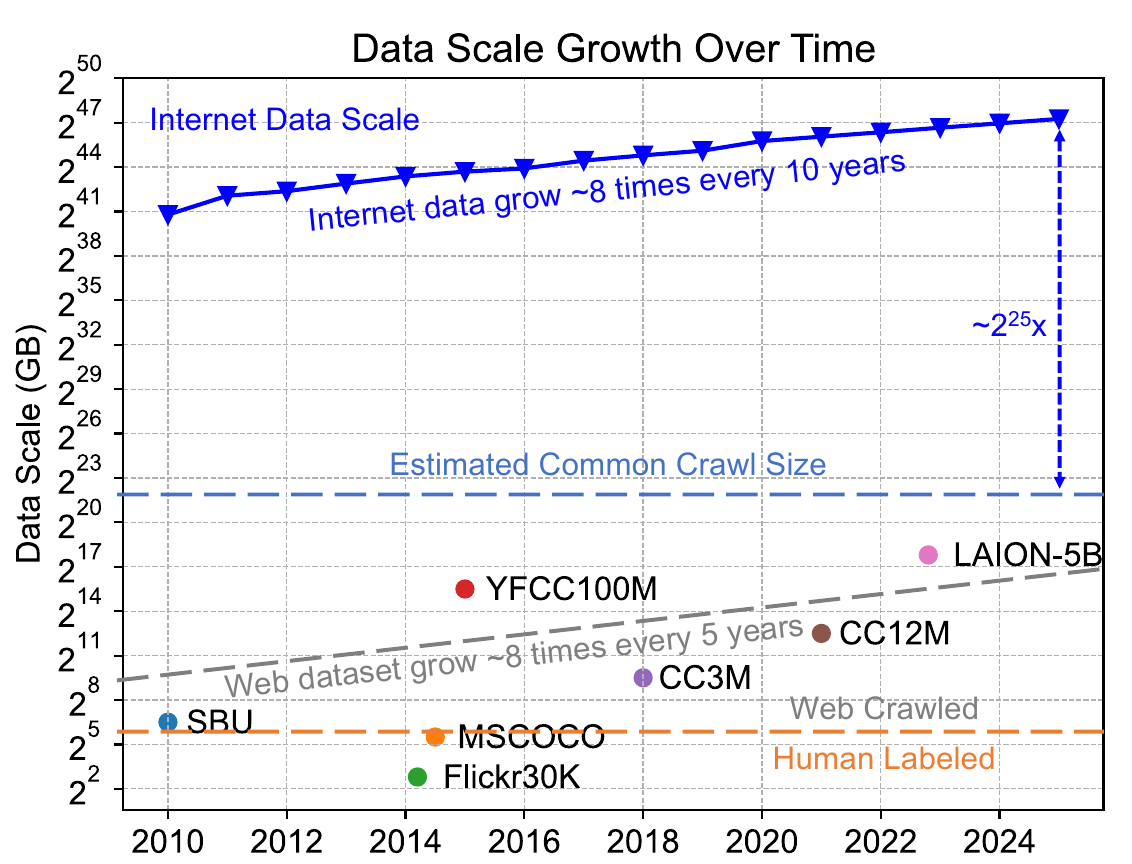}
\caption{Web data grow exponentially with only a small fraction used for deep learning. Human labeling/cleaning is no longer feasible for web-scale datasets.}
\label{fig:web_data}
\end{subfigure}
\begin{subfigure}{0.48\textwidth}
\includegraphics[width=\linewidth]{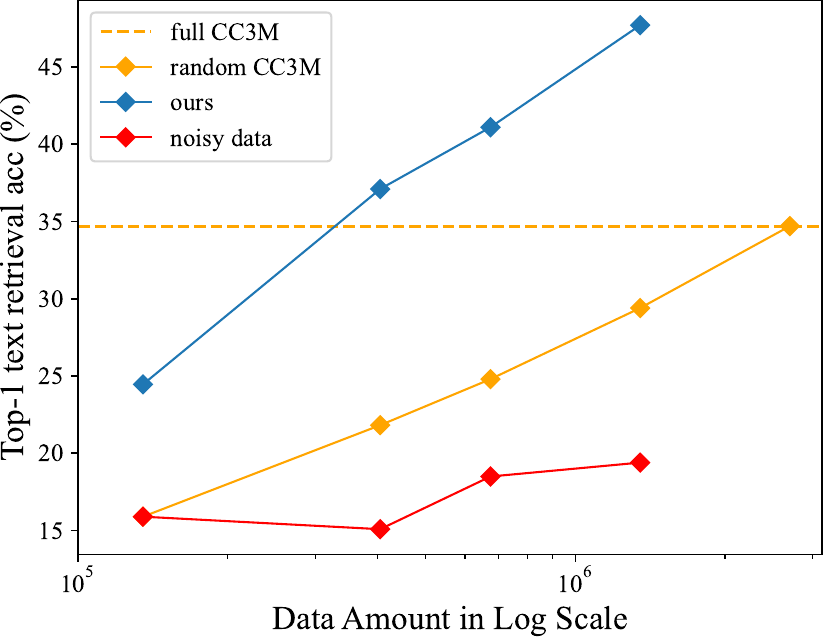}
\caption{Data cleanness is important to training. Better processing can lead to much higher performance at the same data scale.}
\label{fig:different_data_quality}
\end{subfigure}

\caption{Data grows rapidly on an exponential scale. A dataset with cleanness and diversity leads to better data efficiency and training results.} 
\label{fig:motivation}
\vspace{-10pt}
\end{figure*}

Deep learning benefits from the growing abundance of available data. Various datasets~\cite{lecun-mnisthandwrittendigit-2010, cifar,imagenet,jft,sharma-etal-2018-conceptual,commoncrawl} have been created and shared, contributing to the development of deep learning applications. In general, the scale of available data keeps growing (Fig.~\ref{fig:web_data} web data growth).
Even for a single task, there would be multiple datasets with a growing size over time. It has been almost impractical to manually maintain nowadays datasets.
How to efficiently deal with the growing scale of data is becoming a challenge.

Specifically, many prominent datasets~\cite{lin2014microsoft, imagenet} were collected from the Internet and then cleaned up by human annotators at a high cost. However, data continues to emerge on the Internet. According to~\cite{Duarte_2023}, 0.33 zettabytes of data will be created every day in 2024 with an increasing speed. Under these circumstances, collecting web data and doing traditional human cleaning is no longer feasible.

Web data usually has complex sources and various qualities, so directly adding them to the training would reduce the benefit of already collected high-quality data, as web data are likely to have noise and redundancy. 
The problem of how to efficiently utilize web data is worth studying. BLIP~\cite{li2022blip} adopts bootstrapping to filter and relabel noisy web data based on the model's self-prediction and human-annotated finetuning.
Its remarkable results suggest that their data processing greatly improves data quality.
However, BLIP involves model training on all collected data, which may be difficult to scale to even larger data scales.

To scale data processing to an even larger scale, we define the problem in a recursive way. Given a set of data already collected, a new data point can be divided into 3 types: (1) out-of-distribution (noisy) data, (2) redundant data, and (3) clean data with new information. Noisy data need to be filtered out or relabeled; redundant data provide marginal benefit compared to clean data with new information, so its sampling frequency should be correspondingly lowered. Following the above analysis, creating a good dataset can be interpreted as repeatedly adding one informative sample to an already maintained dataset.

Based on this idea, we propose InfoGrowth, a novel framework for efficient online dataset growth with cleanness and diversity awareness. It takes the streaming data as input and returns a clean and diverse dataset. Specifically, for adding a new data point, we estimate its noisy degree and redundancy according to its neighborhood in the already collected dataset. Noisy samples are then relabeled, whereas redundant samples are sampled less frequently or removed. The near-neighbor search is based on an efficient online algorithm, so InfoGrowth has better efficiency and scalability than the previous dataset cleaning algorithms.

InfoGrowth works for both single-modality tasks and multi-modality tasks across different model architectures. It can achieve 2x$\sim$4x data efficiency in both settings, even surpassing its offline counterparts. It is designed to be efficient and scalable, which would be practical for real-world data engines. We hope this work will benefit the dataset preparation of deep learning and sustainable AI.

\section{Background and Related Works}
\subsection{Deep Learning Datasets} 

Datasets are a crucial part of deep learning, serving as the foundational bedrock upon which deep learning models are trained and validated.
Single-modal datasets emerge since the beginning stage of deep learning. MNIST~\cite{lecun-mnisthandwrittendigit-2010}, CIFAR10/100~\cite{cifar}, ImageNet~\cite{imagenet}, COCO~\cite{lin2014microsoft} are the most classical datasets for computer vision, which were all human-labeled, with size ranging from $\sim$10k to 10M. Later datasets with much larger sizes like JFT~\cite{jft} are labeled automatically by algorithm and have a size from 300M to billion, which is far beyond human capability to label. This transition from human labeling to algorithm labeling also happens in other fields like multi-modal datasets.

Multi-modal Datasets encompass diverse data types such as images, text, speech, and audio with paired relations. They can capture rich and nuanced information.
COCO~\cite{lin2014microsoft}, Flickr30K~\cite{young-etal-2014-image}, and Visual Genome~\cite{krishna2016visual} are human-labeled datasets with high-quality image text data pairs. SBU~\cite{sbu} collects data from the internet and does a cleaning to select useful pairs. 
YFCC100M~\cite{Thomee_2016} filters 100 million media objects from Flickr with certain criteria, with metadata paired. 
CC3M~\cite{sharma-etal-2018-conceptual} extract and filter image caption annotations from billions of web pages using Google APIs; with the same procedure and relaxed constraints, CC12M~\cite{DBLP:journals/corr/abs-2102-08981} was created. 
LAION~\cite{schuhmann2022laion5b} filter data from Common Crawl~\cite{commoncrawl} with CLIP~\cite{radford2021learning} score, resulting in a 400M and 5B dataset respectively. WebVid~\cite{Bain21} collects 10 million video clips with captions from the internet. We can see when the dataset size comes to hundreds of millions, human cleaning/labeling is almost infeasible, and automatic methods are employed to craft most extra-large scale datasets.
Meanwhile, extra-large scale datasets like LAION-5B remain almost uncurated, let alone Common Crawl~\cite{commoncrawl} which grows in billion pages monthly.

\subsection{Web Scale Data Curation} 
To reduce the influence of noise in web data, researchers have proposed many techniques. For multimodal datasets, the methods mainly include soft-labeling, filtering, and recaptioning. ALBEF~\cite{albef} utilizes a momentum model to provide pseudo-targets and take weighted mean with the original target, serving as denoised soft labels. BLIP~\cite{li2022blip} follows the design of ALBEF and further cleans the dataset with captioning and filtering using the same model trained on noisy data and fine-tuned on clean data. It has been shown that much higher data efficiency can be achieved with cleaned data. 

For single modal data, DivideMix~\cite{li2020dividemix} and Segment Anything~\cite{kirillov2023segany} are two representative works. DivideMix employs semi-supervised training to fight against noise, training two models together that supervise each other's data cleanness and detect the other's training data noise. In contrast, Segment Anything uses model-assisted human annotation together with model-labeled data to build a semi-supervised training cycle, aligning the model predictions to human annotations. The majority of these works involve model training on given data and/or human annotation. Segment Anything even train the model for multiple runs.

\section{How to Make a Dataset Grow}
\subsection{Preliminaries}
\textbf{Multimodal Models.} CLIP~\cite{radford2021learning} is an influential multimodal model that employs natural language supervision for classification tasks. It trains one image encoder and one text encoder with contrastive loss to predict the image-text pairwise relationship, which greatly improves training efficiency compared to traditional supervised learning using labels.
Following it, BLIP~\cite{li2022blip} further adds two other tasks together for training: predicting whether a chosen image-text pair is matching (image-text matching loss), and predicting masked text based on the image (language modeling loss).
Another improvement is its soft labeling from ALBEF\cite{albef}. These modifications lead to a better capability of BLIP than CLIP, allowing it to perform more vision language tasks (such as captioning, VQA) other than classification or retrieval. More recently, BLIP-V2~\cite{li2023blip2} utilizes Q-Former to bridge the representation of trained image encoder and large language model. MiniGPT-4~\cite{zhu2023minigpt4} further adds an instruction fine-tuning stage to BLIP-V2 and uses Vicuna~\cite{peng2023instruction} as the language model. These modifications equip multimodal models with stronger language skills.

\textbf{Submodular Function} is a set function that describes the decreasing benefit of adding one element to a set. Let $\Omega$ be a finite set, submodular function $\textit{f}$ is defined as:

\begin{equation}
    \begin{aligned}
    &\textit{f}: 2^\Omega \rightarrow \mathbb{R}, \forall x \in \Omega,
    \forall X,Y \subseteq \Omega\ with\ X \subseteq Y,  \\
    & \textit{f}(X\cup\{x\}) - \textit{f}(X)\geq \textit{f}(Y\cup\{x\})- \textit{f}(Y).
    \label{eq:submodular}
\end{aligned}
\end{equation}

This kind of function is broadly used in many fields. We design a function inspired by it to efficiently estimate the information gain of selected samples.

\textbf{Online Approximate Near Neighbor Search Algorithm.} Traditional submodular function cost $O(n)$ for each element. This would sum to $O(n^2)$ overall cost, which is not practical for today's billion-scale dataset. Therefore to reduce this cost, we introduce online approximate near neighbor search (e.g. HNSW~\cite{DBLP:journals/corr/MalkovY16}) to do an efficient neighbor search in $O(logn)$ time and then calculate a neighborhood-based gain. The submodular property~\ref{eq:submodular} still holds, while neighborhood samples provide a faster estimation of the gain. In even larger scale settings, distributed approximate near neighbor search~\cite{douze2024faiss} could be employed.

\subsection{Threotical Analysis}
\label{TA}

In general, machine learning datasets can be interpreted as conditioned data. Despite there are various types of datasets, it is possible to interpret all datasets from the same Bayesian inference perspective. For example, classification tasks learn the conditioning of image data on class; multimodal retrieval tasks learn the mutual conditioning of data pair; auto-regressive next token prediction learns the next token distribution conditioned on the previous context. 
Most datasets are used to capture a conditioned distribution $P(d|c)$ or $P(c|d)$ where d is the data and c is the interested condition. 

For classification tasks, the interested condition c is the class of given data, and the target is $P(c|d)$
. Multimodal retrieval tasks are to predict $P(d_1|d_2)$ and $P(d_2|d_1)$ where $d_1,d_2$ are paired data in different modalities; their corresponding condition is each other. Unsupervised datasets are also conditioned on themselves, at either a coarse-grained level (sample-wise, such as contrastive learning) or a fine-grained level (pixel-wise or token-wise, such as MAE and Next Token Prediction). 
Their intrinsic differences are in the mapping relationships: many-to-one, many-to-many, one-to-many. From this angle of view, datasets of different types can be unified.

Leveraging this insight, we can first map data into an embedding space where the embeddings are separable according to conditions c, then we can detect the out-of-distribution samples (noise) and redundant samples (low information gain for the task) in the embedding space. A good multimodal model would be a preferred encoder to provide a unified view. Its training objective targets to make the embeddings distributed separably in the space according to natural language. If a multimodal model is strong enough, all data would be theoretically hyper-surface separable by human-defined conditions as long as the condition can be defined by natural language.

Within an established embedding space, given a set of correctly conditioned data, we can estimate the quality of a new given data point. By adding good data points step by step and efficiently scaling this assessment, it is possible to make datasets grow, with a higher efficiency than ``collect then clean''.

\subsection{Overview of InfoGrowth}
\begin{figure*}[t]
\centering
\includegraphics[width=1.0\linewidth]{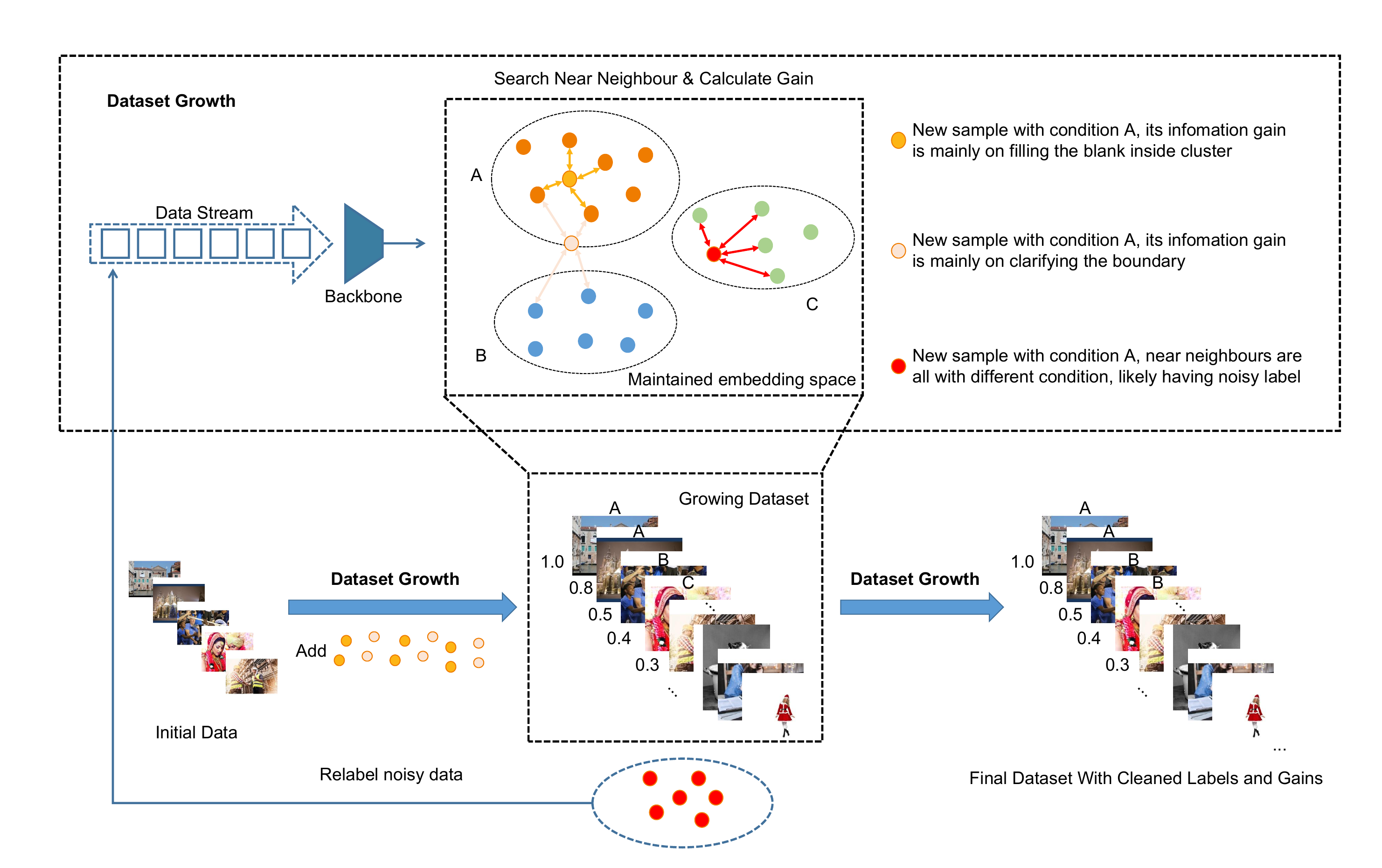}
\vspace{-10pt}
\caption{Pipeline of InfoGrowth. Streaming data first goes through cleaner, then gain calculator, and finally selector.} 
\label{fig:pipeline}
\vspace{-10pt}
\end{figure*}

Following the theoretical analysis, we design our method InfoGrowth as shown in Fig.~\ref{fig:pipeline}. Given an input data stream, each data point is first encoded by a backbone model (we use BLIP here) to be an embedding. Then in this embedding space, we design an efficient way to estimate the quality of the data point according to already cleaned data. Clean ones are added to the dataset with calculated "gain", while noisy ones are relabeled and re-estimated. In short, the pipeline mainly consists of: a multimodal encoder model as the backbone, a cleaner for data cleaning, a gain calculator to measure the information gain, and a sampler for choosing samples. The algorithm can be summarized as Alg.~\ref{alg:pipeline}.
\renewcommand{\algorithmicrequire}{\textbf{Input:}}
\renewcommand{\algorithmicensure}{\textbf{Output:}}
\begin{algorithm}\caption{InfoGrowth Algorithm (for 1 step)}\label{alg:pipeline}
\begin{algorithmic}
\Require Streaming\ data\ pairs\ $(d_i,c_i)$, neighbourhood\ hyperparameter\ k, list $L$, NearNeighbour\ search\ architecture $NN$
\Ensure Processed\ data\ list\ $L$\ with\ gain
\State $p \gets Cleaner(c_i, d_i)$
\If{$p < \delta$}
    \State $c_i \gets Cleaner(d_i)$
    \State $p \gets Cleaner(c_i, d_i)$
    \If{$p < \delta$}
        \State Continue
    \EndIf
\EndIf
\State $d_{feature} \gets Encoder(d_i)$
\State $Neighbors=NN(d_{feature},k)$
\State $g_i \gets mean(cosine\text{-}distance(d_{feature},Neighbours))$
\State Add $(d_i,c_i, g_i)$ to L, sample based on $g_i$ values
\State Add $d_{feature}$ to $NN$
\end{algorithmic}
\end{algorithm}
\vspace{-10pt}

\subsection{How to detect and correct noisy sample}

Cleaner, by its nature, is a model predicting $P(c|d)$. When the cleaner prediction has a lower noise than the given data, its prediction could help improve the label quality. Following BLIP~\cite{li2022blip}, on multimodal datasets, the cleaning process mainly uses filter and recaption to improve the data alignment and reduce noise. We use a trained BLIP model (vit-base-129M) as the encoder and encode image/text into the multimodal representation. The recaptioner we use is MiniGPT-4. Benefiting from the latest development in LLM and multimodal training, it can provide better recaption with a comprehensive understanding of the images. For single modality tasks, a cleaner would be a trained model predicting $P(c|d)$ so that it can detect suspicious samples and relabel the sample. 
A difference in our procedure is that we use the cosine similarity of two modalities' encoded features to filter samples whose multimodal cosine similarity is smaller than $\delta$. This avoids the overhead of using another encoder in BLIP to identify unmatched samples. 
The $\delta$ value can be decided in two ways: use a predefined value or update it with online data statistics. 

\subsection{Gain Calculation}
\label{sec:gain}

Inspired by submodular functions, we design a submodular-like function for calculating the information gain of a data pair $(d,c)$.
Traditional submodular functions take $O(n)$ time to calculate a sample's information gain, which would be a huge overhead for growing billion-scale datasets. 
To efficiently calculate the gain, we exploit an online approximate nearest neighbor search architecture HNSW~\cite{DBLP:journals/corr/MalkovY16} to query the nearest k neighbors of a given data point. The cost of query and update are both $O(log(n))$ where $n$ is the number of collected data points. Then we calculate the cosine distance of those neighbors to the given data, using the mean of distance as the information gain: $Gain_{Info} = mean_i([cos\text{-}dis(d,neighbour_i)])$. This design provides a submodular-like property ($U \subseteq V \Rightarrow f(x,U)\geq f(x,V)$) with higher efficiency ($O(logn)$ compared to set operation O(n)). It also has a good awareness of the neighborhood to avoid redundancy. We then use the gain values for sampling, which further improves the data efficiency beyond cleaning.

On the classification task, we add an "entropy gain" into consideration. It is defined to reflect how much a new sample helps to clarify classification boundary: $Gain_{Entropy}=1-p$, where $p$ is the probability of using the near neighbor's class label to correctly predict the given data's label. This gain puts more weight on boundary data points than center data points.

\subsection{How to select informative samples}
We design two kinds of samplers: static and dynamic. The static one aims to get better diversity with a fixed total amount of data, while the dynamic one aims to get better computational efficiency on more collected data. Our static sampling sets the probability of choosing a sample as:
\begin{equation}
    \mathcal{P}(x_i)  = \frac{G_i}{\sum_{j=1}^{|L|}G_j}.
\label{eq:static_sampling}
\end{equation}
Based on the probability given by Eq.~\eqref{eq:static_sampling}, we randomly choose samples without replacement to the target number. Our dynamic sampling uses a two-phase sampling strategy, with one phase for overall diversity and another phase for generalization (local diversity). The first phase has the same probability as Eq. \eqref{eq:static_sampling}, and use a total amount of $\sum_{j=1}^{|L|}G_j$; the second phase use the following probability:
\begin{equation}
    G'(x_i)  = max(0.1, 1 - G_i),
    \mathcal{P}(x_i)  = \frac{G'_i}{\sum_{j=1}^{|L|}G'_j},
\end{equation}
and sample a total amount of $\lfloor\sum_{j=1}^{|L|}G'_j\rfloor$. In the training, the two phases take turns alternately at each epoch. This leads to a cost between 50\% and 55\% of training with access to all data.

\section{Experiments}
We verify the effectiveness of our method on multiple popular datasets with different architectures and tasks, including 1. BLIP pretraining on multimodal dataset CC3M and retrieval/caption evaluation. 2. ResNet-50 classification on single-modal dataset ImageNet-1K~\cite{imagenet}

\subsection{Datasets and Implementation Details}

\textbf{CC3M} is a dataset with web-crawled image-text pairs, filtered with Google API keeping only 0.1\% of the raw data. This dataset contains a total of
2.7 million images. We use it to demonstrate the effect of our algorithm in different situations. It is the default dataset for ablation experiments. 

\textbf{ImageNet} is a dataset with web-crawled images and human-annotated labels across a large number of classes. We use ImageNet-1K which contains 1000 classes and 1.28M images to evaluate our method on single-modality.

\textbf{Implementation Details}. For backbone encoding, the process is parallelizable without constraint on accelerator number. CC3M cleaner uses MiniGPT-4l which requires at least 1 NVIDIA A100 GPU. Other parts of the pipeline (gain calculation and sampling) can be run on the CPU.
For the training of BLIP, we use PyTorch implementation from the official version and train it on 8 NVIDIA A100 GPUs. The model is
pre-trained for 20 epochs with a batch size of 640 and an
AdamW [27] optimizer with a weight decay of 0.05. We follow the default learning rate warm-up from 1e-6 to 3e-4 and a linear decay with a rate of 0.9. For image data augmentation, we utilize RandAugment [8] except for color inversion. During pre-training, we randomly crop images to a resolution of 224 × 224. Other implementation details can be found in the supplementary material.

\subsection{Evaluation on Multimodal Dataset}

\textbf{Image-text retrieval} consists of two parts: (1) image as the query to retrieve texts (TR); (2) text as the query to retrieve images (IR). The pre-trained model is evaluated on MSCOCO \cite{lin2014microsoft} and Flickr30K~\cite{young-etal-2014-image}. For the COCO zero-shot setting, the pre-trained model is directly evaluated on the test data. For zero-shot retrieval on Flickr30K, we follow the procedure proposed in~\cite{li2022blip} (zero-shot evaluation on Flickr with the model fine-tuned using MSCOCO). For the fine-tuning setting, the model is further fine-tuned on the  Flickr30K training data and evaluated on the validation/test data. 

\begin{table*}
\centering
\caption{\textbf{Fine-tuning and \colorbox{bluebell}{zero-shot} image-text retrieval} results on MSCOCO and Flickr30K. InfoGrowth demonstrates an improved data efficiency.}
\footnotesize
\resizebox{\textwidth}{20mm}{
\begin{NiceTabularX}{\linewidth}{y{28}ly{28}|cccccc|cccccc}
\CodeBefore
  \rowlistcolors{4}{=,=,=,=,blue!10,=,=,=}[restart,,cols={2-15}]
  \Body
    Model & Dataset & Size  & \multicolumn{6}{c}{ MSCOCO (5K test set)}  & \multicolumn{6}{c}{ Flickr30K (1K test set)}  \\
      &&&\multicolumn{3}{c}{Image$\rightarrow$ Text}&\multicolumn{3}{c}{Text$\rightarrow$ Image}&\multicolumn{3}{c}{Image$\rightarrow$ Text}&\multicolumn{3}{c}{Text$\rightarrow$ Image}\\ &&&R@1&R@5&R@10&R@1&R@5&R@10&R@1&R@5&R@10&R@1&R@5&R@10\\
\shline

\multirow{8}{*}{BLIP} 
    & \CCThree  &2.71M  & 70.9 & 90.6 & 95.3 & 54.3 & 79.8 & 87.4 & 89.2 & 98.4 & 99.6 & 75.1 & 93.2 & 96.4 \\
      & Ours &0.40M  & \bf 70.9  &  \bf91.3  & \bf 95.9  &  53.3 &  79.5  & 87.0  & 84.6  & 97.1  &  98.7  & 69.5   &  90.3 & 94.5   \\
       & Ours&0.68M  &   \bf71.5  &  \bf91.1  & \bf95.3  &  \bf53.9  & \bf79.5   & \bf87.5 & 89.0   &  98.4  & 99.0  &  74.6  & 92.9  &  95.9   \\
       & Ours&1.35M  &   \bf72.1  &  \bf92.2  & \bf 96.1  &  \bf55.3  &  \bf81.4  & \bf89.0    & \bf91.5  &  \bf99.1  & \bf99.8  & \bf77.7   &  \bf94.0  & \bf96.9    \\
      & \CCThree &2.71M &   34.7 & 60.7  & 71.9 & 28.0 & 52.5 & 63.7 & 87.4 & 98.2 & 99.5 & 72.8 & 90.7 & 94.6  \\
      & Ours & 0.40M   & \bf37.1  &  \bf63.8  & \bf74.2   &  \bf28.1  &  \bf53.8  & \bf65.1  & 82.2   &  96.9  & 98.6  &  67.0  & 87.8   &  92.7 \\
       & Ours &0.68M   &\bf41.1  &  \bf67.7  & \bf77.5   &  \bf32.8  &  \bf58.2  & \bf68.8   &  86.8  & 97.7  &  99.0  & 70.2   &  90.2 &90.4    \\
       & Ours &1.35M   &\bf47.7  &  \bf73.0  & \bf82.0   &  \bf38.1  &  \bf63.3  & \bf73.2   & \bf 88.7  &  \bf98.3  &99.0   &  \bf74.2  & \bf92.2   &  \bf95.4  \\
\midrule
\multirow{2}{*}{CLIP} 
    & \CCThree &2.71M  & 60.4 & 85.3 & 93.2 & \bf 48.9 & 75.4 & 84.7 & 77.3 & 91.1 &  \bf 93.2 & 71.6 & 90.1 & 91.4 \\
      & Ours&0.68M  & \bf 62.2 & \bf 86.4 & \bf 94.0 & 48.6 & \bf 76.3 & \bf 86.0 & \bf 80.5 & \bf 94.6 &  \bf 98.3 & \bf 72.3 & \bf 90.6 & \bf 93.0  \\
      
\end{NiceTabularX}
}

\label{tab:ft_and_zero_shot_retrieval}
\vspace{-20pt}
\end{table*}

\begin{figure*}[h]
    \centering
    \begin{subfigure}{0.45\linewidth}      {\includegraphics[width=\linewidth]{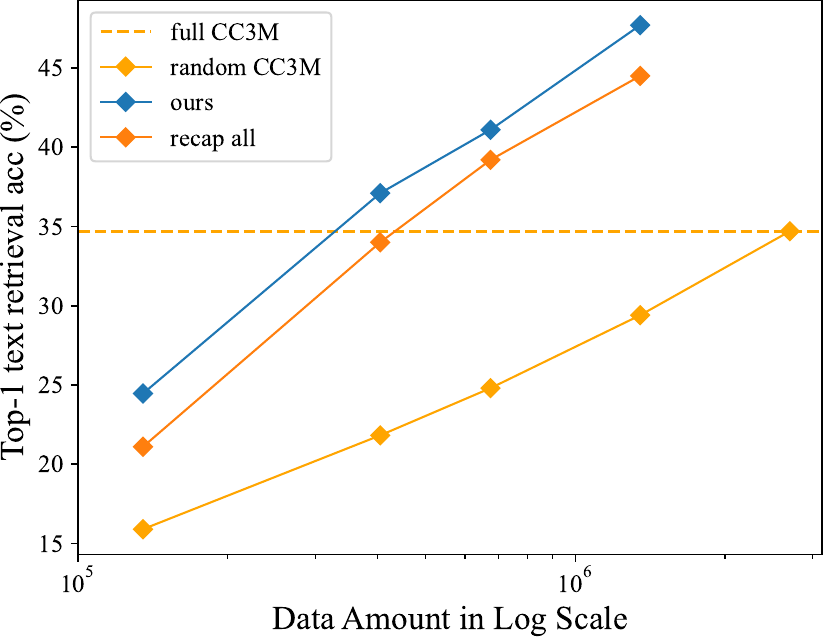}}
      \caption{Text retrieval performance.}
      \label{fig:tr_data_efficiency}
    \end{subfigure}
\begin{subfigure}{0.45\linewidth}
\includegraphics[width=\linewidth]{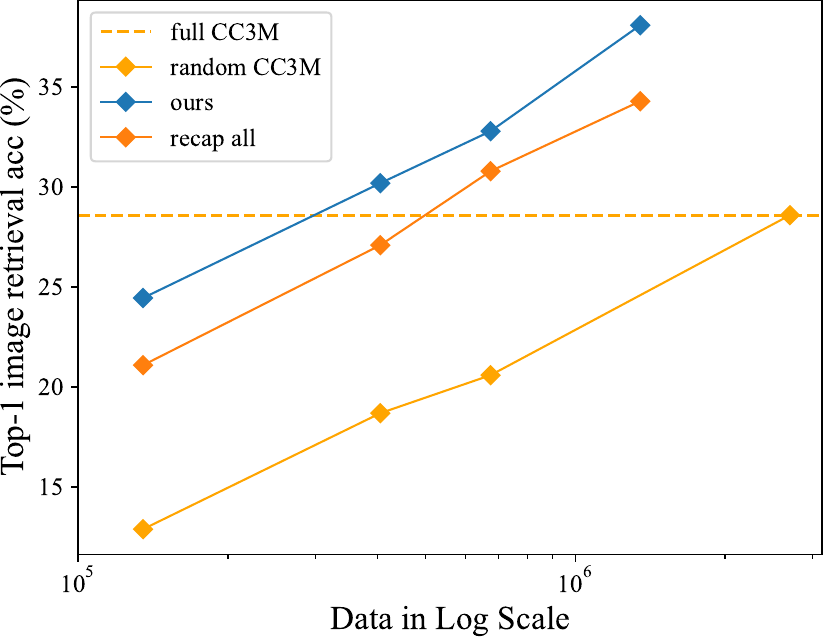}
    \caption{Image retrieval performance.}
    \label{fig:ir_data_eficiency}
    \end{subfigure}
\caption{InfoGrowth demonstrates better training results than using original CC3M and MiniGPT4 recaptioned CC3M at the same amount of data. Its cost is also much lower than recaptioning all data.}
\label{fig:data_efficiency}
\vspace{-30pt}
\end{figure*}

As seen in Fig.~\ref{fig:data_efficiency}, we train the BLIP model on the dataset processed by our method and the original CC3M. Our processed dataset uses less data and has a much higher data efficiency than using the original CC3M. With only 0.4M data (0.8M seen), our static dataset can surpass the training performance using the full original CC3M. Moreover, with our gain-based dynamic two-phase sampling (saving at least 45\% training cost), the training takes only 8.2\% of the CC3M's computation and achieves a 3 percent zero-shot text retrieval performance improvement on COCO. When the sample number increases, this gap is further enlarged. Compared to only recaptioning samples, InfoGrowth shows higher data efficiency with gain-based sampling. 
In Table~\ref{tab:ft_and_zero_shot_retrieval}, we can see the dataset processed by InfoGrowth demonstrates a strong zero-shot transfer ability on MSCOCO, using less than $1/6$ data to surpass original CC3M. On Flickr30K, the data needed is increased to $1/2$. 

\vspace{-20pt}

\begin{table*}[h]
	\centering
 \caption{
        \textbf{Comparison with BLIP model pre-trained on different data sources} for VQA, NLVR$^2$, RefCOCO+ and COCO Captioning. 
        CLIP architecture can not be evaluated on part of these tasks due to structural limitations.
	}
 \vspace{-5pt}
 \resizebox{\columnwidth}{!}{
 \footnotesize
	\begin{tabular}	{ll|lllllllll}
	 \multirow[t]{2}{*}{Dataset}& \multirow[t]{2}{*}{ \#Samples}& \multicolumn{2}{c}{VQA} & \multicolumn{2}{c}{NLVR$^2$} &\multicolumn{2}{c}{COCO Caption} &\multicolumn{2}{c}{Nocaps Validation}  \\
	  && test-dev & test-std & dev & test-P & B@4 & CIDEr  & CIDEr & SPICE  \\
	  		\shline 	
    \demph{Random-\CCThree} & \demph{0.67M} & \demph{68.3} & \demph{66.2} & \demph{73.6} & \demph{73.8} &\demph{35.9} &\demph{118.1}&\demph{85.3}&\demph{11.6} \\
	   \CCThree~\cite{sharma-etal-2018-conceptual}&2.7M & 71.3  & 71.4  & 75.9 & 76.5  & 36.1 & 121.4 & 90.6 & 12.8 \\
        Ours & 0.67M & 70.9$_{{-0.4}}$  & 71.0$_{{-0.4}}$  & 75.4$_{-0.5}$  &   76.8$_{\hl{+0.3}}$ & 36.7$_{\hl{+0.6}}$  & 121.7$_{\hl{+0.3}}$ & 89.8$_{{-0.8}}$ &  13.1$_{\hl{+0.3}}$ \\
        Ours & 1.35M & 71.6$_{\hl{+0.3}}$  & 71.9$_{\hl{+0.5}}$  & 76.1$_{\hl{+0.2}}$  &   77.4$_{\hl{+0.9}}$ & 36.9$_{\hl{+0.8}}$  & 122.3$_{\hl{+0.9}}$ & 90.6$_{{+0.0}}$ &  13.3$_{\hl{+0.5}}$
	\end{tabular} 
 }
 \label{tab:vqa_nlvr}
    \vspace{-20pt}
\end{table*}

\subsubsection{Downstream Tasks} 
In the next paragraphs, we evaluate our trained model on downstream cross-modal tasks. We first introduce these tasks:

\textbf{Visual Question Answering (VQA.)} We evaluate our
model’s performance on the VQA~\cite{goyal2017making} task, where the model
needs to provide an answer based on an image and a question. We consider it as an answer generation task that allows
open-vocabulary VQA for better results, following previous
works [22, 23]. The results are presented in Table~\ref{tab:vqa_nlvr}. The
BLIP trained on InfoGrowth outperforms baseline by
InfoGrowthx\% on test-dev splits, demonstrating the effectiveness of
our InfoGrowthx for improving VQA performance.

\textbf{Visual Reasoning.} The Natural Language Visual Reasoning task is a binary classification task that requires the model to reason about two images and a question
in natural language. The dataset ($NLVR^{2}$)~\cite{suhr2019corpus} contains 107,292 examples of human-written English sentences paired with web photographs. Multi-modal reasoning is crucial for
the completion of this task. We observe that BLIP trained
on our dataset achieved InfoGrowth\% accuracy compared to InfoGrowthx\%
achieved by the CC3M, as shown in Table~\ref{tab:vqa_nlvr}.

\textbf{Image captioning.} The task involves describing an input image, which we evaluate using No-Caps and COCO
datasets. Both datasets are fine-tuned on COCO with the
Language Modeling (LM) loss. We adopt a zero-shot setting for No-Caps dataset, and start each caption with the
phrase “a picture of” for the BLIP architecture. We do not
pre-train using COCO to avoid information leakage. Our
results outperform the baseline with a much smaller quantity of
pre-training data, as shown in Table~\ref{tab:vqa_nlvr}.

\subsection{Evaluation on Single-modal Dataset}
Besides the multimodal datasets, InfoGrowth can also be applied to single-modal datasets as analyzed in Sec.~\ref{TA}. A distribution of $P(d|c)$ can be established in the embedding space, and the target objective $P(c|d)$ can be transferred by Bayesian's rule to $P(d|c)P(c)/P(d)$, which is proportional to $P(d|c)$. Therefore, the corresponding data quality can be estimated within the same framework as in the previous subsection.

\begin{table}[h!]
\vspace{-10pt}
    \centering
    \caption{
    Comparison of performance and time cost on ImageNet-1K. Results are reported with ResNet-50 under 50\% prune ratio for 90 epochs on an 8-A100-GPU server. ``Time" is wall clock time; ``Total (n*h)'' is the total node hour.
    }
    \vspace{-8pt}
    \begin{tabular}{ccccc|c}
    \toprule
    & GC~\cite{iyer2021submodular}      & EL2N-20~\cite{toneva2018empirical}              &   UCB~\cite{raju2021ddp}    & InfoGrowth       & Full Data \\ \midrule
    Acc (\%)    & 72.8$_{\pm0.4}$ & 74.6$_{\pm0.4}$  & 75.3$_{\pm0.3}$ &  \textbf{75.8$_{\pm0.3}$}  & 76.4$_{\pm0.2}$ \\
    \midrule
    Training (h)    &8.75  &8.75   &8.75  &8.75  & 17.5 \\ 
    Overhead (h)     & $>$24    & $>$17.5  & 0.03     &  0.8     & 0.0  \\
    Total (n*h)     & $>$94  & $>$210  & \textbf{70}   & \textbf{70.8} & 140.0  \\
    \bottomrule
    \end{tabular}
    \label{tab:imagenet_comparison}

\end{table} 

\textbf{ImageNet Classification.} We evaluate InfoGrowth on ImageNet-1k~\cite{imagenet} Classification task using ResNet-50~\cite{DBLP:journals/corr/HeZRS15}. The corresponding gain is defined in Sec.~\ref{sec:gain} as the mean of information gain and entropy gain. The corresponding result is shown in Table~\ref{tab:imagenet_comparison}. One can find that InfoGrowth not only significantly surpasses traditional static dataset compression methods like GC and EL2N in both performance and computation cost, but also outperform dynamic pruning methods like UCB in final accuracy. These results are suggesting InfoGrowth's generality and competitiveness across task, data, and model architecture.

\subsection{Ablation}
If not stated otherwise, we train BLIP with 0.4M CC3M data processed with the specified ablation setting and all other settings as default, and evaluate on the COCO zero-shot retrieval task.

\textbf{Evaluating the components of InfoGrowth.}
We design an ablation study to investigate the characteristics of cleaner and sampler modules in table~\ref{tab:abl_module}. We train BLIP on the dataset processed by InfoGrowth with different components ablated and evaluate its zero-shot performance on the COCO retrieval task. The baseline is a random selection from CC3M. We can see each module provides a performance gain compared to the baseline, and using caption+filtering cleaning together with gain-based sampling achieves the best performance in our experiment. With only 0.4M data ever seen, BLIP trained on InfoGrowth's processed dataset can achieve comparable retrieval performance with using the full CC3M. Using dynamic sampling, the training cost can be further saved by nearly half, taking only 8\% of the cost on full CC3M.
\begin{table}[h]
\vspace{-10pt}
\parbox{.51\linewidth}{
    \centering
        \caption{\textbf{Component ablation}.  
Both the cleaning and gain-based sampling operations are important to the downstream retrieval. With only 0.4M data seen, InfoGrowth is able to beat full CC3M(2.7M).}
    \begin{tabular}{x{40}x{35}x{35}x{20}x{20}}
    recaption & sampling & time(h) & TR@1&  IR@1 \\
    \shline
    & & 4.2 & \baseline{21.8}  & \baseline{18.7} \\
    \checkmark& &4.2 & 34.1  & 27.1 \\
    &\checkmark & 2.3 & 23.9 & 21.0  \\
    \checkmark&\checkmark & 2.3 & \bf 37.1 & \bf 28.1  \\
    \hline
    \multicolumn{2}{c}{Full CC3M} & 14 & 34.7 & 28.0
    \end{tabular}
    \label{tab:compoent_ablation}

    \label{tab:abl_module}
}
\hfill
\parbox{.47\linewidth}{
 \centering
    \caption{\textbf{k neighbor and Mean method}. Comparison of neighborhood size and arithmetic/harmonic mean to calculate gain. k=4 with mean performing the best in the experiment.}
    \begin{tabular}{y{30}x{25}x{25}x{25}x{25}}
        & \multicolumn{2}{c}{arithmetic} & \multicolumn{2}{c}{harmonic} \\
        \toprule
       k & TR@1 & IR@1 & TR@1 & IR@1 \\
        \shline
        k=1 & 33.5 & 26.7 & 33.5 & 26.7\\
        k=2 & 33.5  & 27.2 & 34.6 & 27.2\\
        k=4 & \baseline{\bf 37.1} & \baseline{\bf 28.1} & 35.3 & 27.7\\
        k=8 & 33.6 & 26.7 & 33.2 & 27.1 \\
    \end{tabular}

    \label{tab:ablation_k}
}
\vspace{-10pt}
\end{table}

\textbf{Influence of neighborhood query number k and way of calculating mean.}
In the gain calculator, we use the neighborhood of a given feature to calculate its information gain. Here we conduct an ablation experiment in Tab.~\ref{tab:ablation_k} to show the influence of different k choices. We train the BLIP model with the dataset processed by different k values. We can see that when using the arithmetic mean, increasing k from 1 to 4 leads to a gradual performance improvement. Further increasing k to 8 shows a performance degradation. When using harmonic mean, more weight is put on closer neighbors, and increasing k gradually improves the performance, as it approximates the submodular gain on full data with higher precision.

\begin{table}[]
\vspace{-15pt}
\parbox{.49\linewidth}{
    \centering
    \caption{\textbf{Gain Function}. Averaged image and text gain function is best for overall performance.}
    \tablestyle{4pt}{1.05}
    \begin{tabular}{y{90}x{20}x{20}}
    case & TR@1&  IR@1 \\
    \shline
    Image Gain & 32.8  & 28.0 \\
    Text Gain & 34.0 & 26.7 \\
    Info+Alignment Gain & 33.7 & 27.2 \\
    Info Gain & \baseline{\bf 37.1 } & \baseline{\bf 28.1 } \\
    \end{tabular}
    \label{tab:abl_gain_func}
}
\hfill
\parbox{.49\linewidth}{
\centering
    \caption{\textbf{Noise ratio}. Our method provides noise resistance.}
    \label{tab:ablation_noise}
    \begin{tabular}{y{45}y{60}x{20}x{20}}
    noise ratio & method & TR@1 & IR@1 \\
    \shline
    0\% &\demph{clean baseline} & \demph{34.7} & \demph{28.0} \\
    \hline
    10\% & unprocessed & 17.8 & 15.2 \\
    10\% & ours & 33.8 & 26.6 \\
    25\% & unprocessed & 15.0 & 13.1 \\
    25\% & ours & 31.2  & 25.6 \\
    \end{tabular}
}
\vspace{-15pt}
\end{table}
\textbf{Multimodal data gain calculation.}
The default multimodal gain (Info Gain) calculation has two parts: image gain and text gain, each calculated with the mean of k nearest neighbors' cosine distance. Alignment Gain is the cosine similarity between two modalities' embedding.
We ablate each part in Tab.~\ref{tab:abl_gain_func} and see their performance difference. We can see that with gain from both modalities, the dataset processed gives the best training result on the retrieval task. Adding Alignment Gain into gain calculation doesn't improve the retrieval performance.

\textbf{Data amount.}
In general, more collected data leads to better training results. From Fig.~\ref{fig:different_data_quality} and~\ref{fig:data_efficiency}, three qualities of data from InfoGrowth, recaption, and original each form a linear line in exponential data growth. However there are some interesting findings to be discussed in Sec.~\ref{sec:marginal_benefit}.

\textbf{Model generalization and Task generalization.}
In Table~\ref{tab:ft_and_zero_shot_retrieval}, we test the performance of InfoGrowth on CLIP architecture. It suggests the effectiveness of growing a dataset within the embedding space of multimodal models is independent to model architecture. 

\begin{wraptable}[9]{r}{0.6\textwidth}
\vspace{-35pt}
    \centering
    \caption
	{Video classification on MSRVTT and Image classification on ImageNet-1K for BLIP.}
\footnotesize
     \begin{tabular}{cc|cccc}
     \toprule
          \multirow{2}{*}{Dataset} &\multirow{2}{*}{Scale}& \multicolumn{3}{c}{MSRVTT} & Imagenet  \\
		 & & R@1$\uparrow$ &  R@5$\uparrow$ &  R@10$\uparrow$ &Acc$\uparrow$\\
         \midrule 
         Rand-cc3m &0.67M& \demph{23.3} & \demph{42.8} & \demph{53.3} & \demph{57.3} \\
        cc3m&2.82M& 26.0 & 46.3 & 58.0 &62.5\\
        Ours&0.67M& 26.4&  47.3&  58.4 &62.1\\
        Ours&1.35M&\bf 28.8& \bf 49.4& \bf 60.0 &\bf62.9\\
        \bottomrule
    \end{tabular}
	\label{tab:infogrowth_video_retrieval}
\end{wraptable}
In Table~\ref{tab:infogrowth_video_retrieval}, we can see that on MSRVTT, which is video classification with different data distribution to BLIP training data, our trained BLIP on 0.67M CC3M data processed by InfoGrowth shows improved ability than trained on the original CC3M. InfoGrowth also improves over the original CC3M on ImageNet-1k zero-shot classification with 1.35M processed data. It suggests that task generalizability is inherited by InfoGrowth processed dataset.

\textbf{Data noisy scale.}
InfoGrowth demonstrates the ability to resist noise as shown in table~\ref{tab:ablation_noise}. When more than 25\% of the captions are randomly shuffled, training directly on it shows a degraded performance. In contrast, InfoGrowth is still able to denoise the data and achieve a reasonable.

\section{Discussion}

\subsection{Comprehensive Comparison with Other Methods}
\begin{table}[h]
    \centering
    \vspace*{-20pt}
\caption{A comprehensive comparison with other methods.}
\label{tab:comparison_methods}
\vspace*{-2pt}
\begin{tabular}{cccccc}
    \toprule
     method & type & functionality  & modality & per sample cost  \\
     \midrule
     InfoGrowth & online & clean and select  & both & Inference, O(logn) growth\\
     Bootstrapping & offline & clean  & multi-modal & Training*2+Inference \\
     Data Pruning & offline & select & single-modal & Inference, O(n) growth \\
     \bottomrule
\end{tabular}
\end{table}
\vspace{-15pt}

InfoGrowth mainly differs from existing dataset processing methods in several aspects: 1. it is an online algorithm 2. it supports data for both single-modal and multimodal tasks 3. it is efficiently scalable to support the data growth.

The cost of InfoGroth is analyzed as follows: $O(M_{encoder})$ per sample to make the decision, $O(kd \cdot logn)$ for search and compare k neighbors with embedding size d, $O(M_{cleaner})$ to clean detected noisy samples ($\sim$30\% samples in CC3M).

We compare the difference of InfoGrowth with other methods in Table~\ref{tab:comparison_methods}. When taking into consideration that data grow on an exponential scale, bootstrapping methods like BLIP could be still acceptable because they produce a model; data pruning methods with $O(n)$ per sample cost would take $O(n^2)$ on an exponentially growing n, and this cost occurs periodically. In contrast, InfoGrowth doesn't incur further costs on already collected data, and deals with the new data in an $O(logn)$ growth. 

A most recent work InfoBatch~\cite{qin2024infobatch} proposes to do unbiased dynamic data pruning to losslessly accelerate deep learning training with negligible cost. It would introduce little cost when more data is added.
However, InfoBatch currently cannot generalize to multimodal tasks because its rescaling doesn't work for contrastive loss, while most multimodal tasks use contrastive learning.

\subsection{Scaling to Even Larger Scale}
Currently, the experiments verify InfoGrowth on a million scale of data using HNSW. To further scale to billion-scale data and real-world data pipeline, InfoGrowth only needs a little modification. Note that data encoding and cleaner prediction could be totally asynchronized. Only the near-neighbor search architecture needs to be synchronized, so a distributed near-neighbor search algorithm~\cite{douze2024faiss} would be enough for further scaling.

\subsection{The Marginal Benefit of Data}
\label{sec:marginal_benefit}
\begin{wrapfigure}{r}{0.32\textwidth}
\vspace{-70pt}
  \begin{center}
    \includegraphics[width=0.32\textwidth]{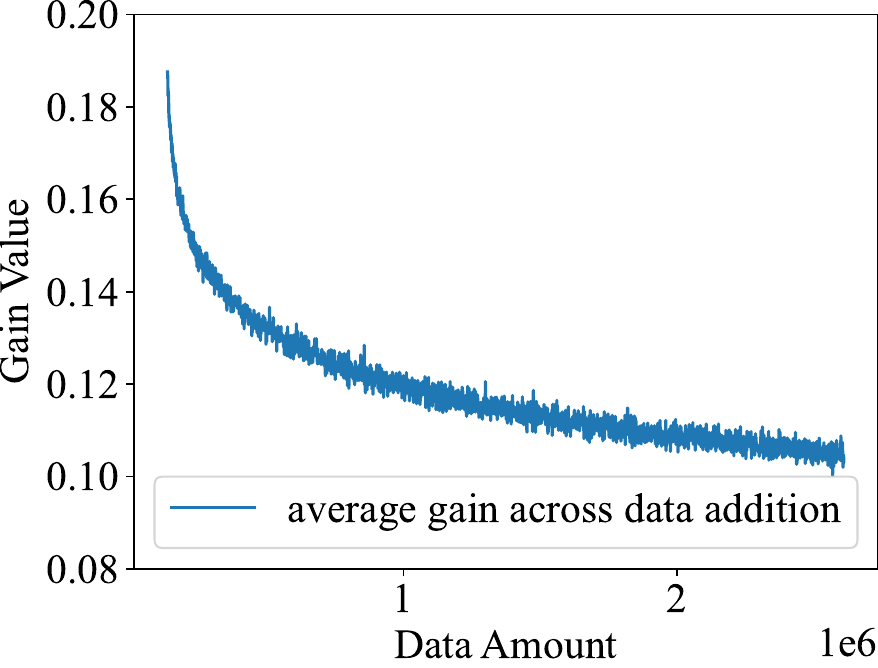}
  \end{center}
  \vspace{-20pt}
  \caption{Gain decay with more data collected.}
    \label{fig:gain_decay}
\vspace{-40pt}
\end{wrapfigure}
From Fig.~\ref{fig:different_data_quality} and~\ref{fig:data_efficiency}, one can see that when data of the same quality is added on an exponential scale (computation increased proportionally), the model performance increases almost linearly. Sun et al.~\cite{jft} reported the same phenomenon which motivated them to create the ultra-large-scale dataset JFT. This also implies that with more data collected, the further gain of added data would decrease. Our defined gain also reflects this trend, which is demonstrated in Fig.~\ref{fig:gain_decay}. 

\begin{wraptable}[6]{r}{0.3\textwidth}
    \centering
    \vspace{-33pt}
    \caption{InfoGrowth with 0.67M selected data.}
    \footnotesize
     \begin{tabular}{c c c}
     \toprule
          Seen Data & TR@1 & IR@1 \\
         \midrule 
         1.35M & 41.1 & 32.8 \\
         2.7M & 40.9 & 32.5 \\
        \bottomrule
    \end{tabular}
    \label{tab:no_growth}
\end{wraptable} 
Another interesting phenomenon of InfoGrowth is that when the total selected data amount is fixed, choosing from more candidate data doesn't improve performance when enough data is seen (Table.~\ref{tab:no_growth}). This implies that the data quality improvement of InfoGrowth has an upper bound because marginal decay.

\subsection{Visualizations}

\begin{figure}
\vspace{-25pt}
    \centering
    \includegraphics[width=\linewidth]{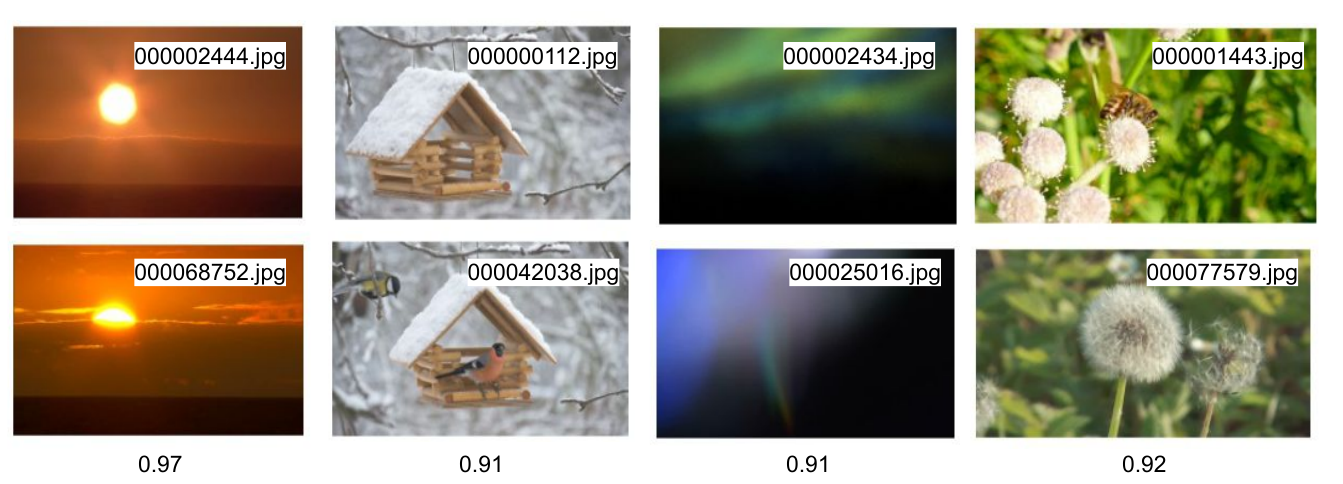}
    \caption{Redundant samples detected in CC3M and their cosine similarity.}
    \label{fig:redundant}
    \vspace{-10pt}
\end{figure}

\begin{wrapfigure}{r}{0.5\textwidth}
\vspace{-35pt}
  \begin{center}
    \includegraphics[width=0.5\textwidth]{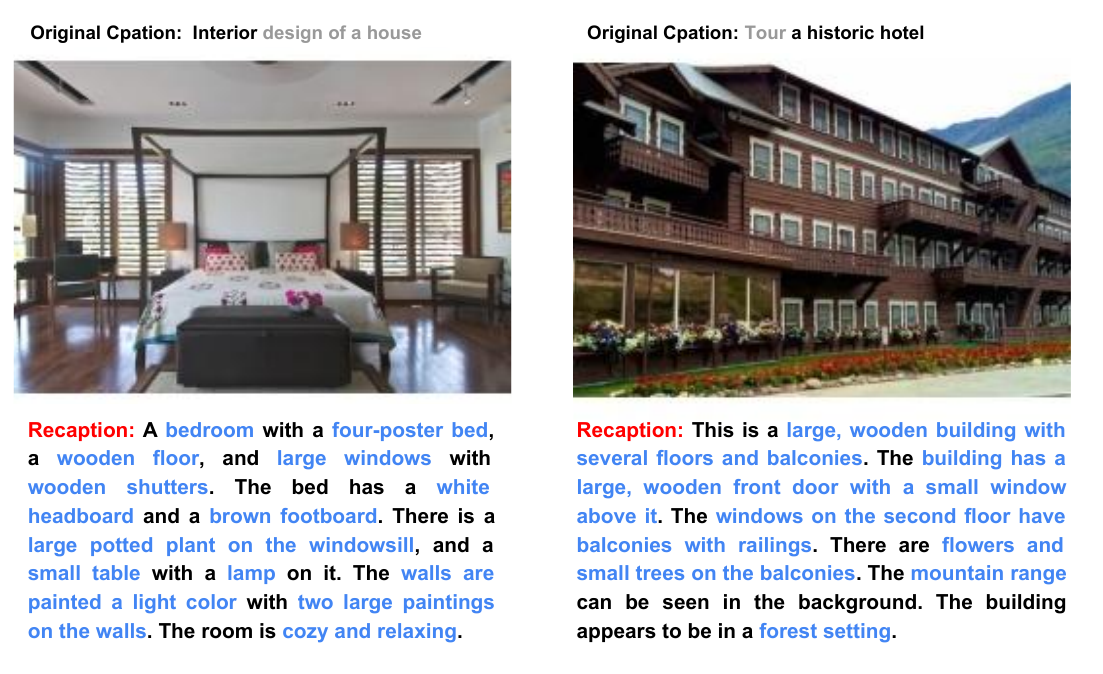}
  \end{center}
  \vspace{-20pt}
  \caption{Cleaner improving data quality. Grey words are low-quality parts and blue words are high-quality descriptions.}
    \label{fig:recap}
\vspace{-60pt}
\end{wrapfigure}
In this section, we visualize in Fig.~\ref{fig:redundant} and ~\ref{fig:recap} some examples from our algorithm's detected redundant/noisy samples. The static selection according to gain and dynamic two-phase sampling both prevent those redundant samples from occurring at the same time. For noisy samples, the cleaner will improve the data quality with recaptioning/relabeling.

\section{Conclusion}
In this work, we propose a novel solution to efficiently deal with the growing data demand of deep learning. By turning the data cleaning and selection process into an online algorithm, our proposed InfoGrowth algorithm demonstrates its competitive capability in improving data quality for both multi-modal tasks and sing-modal tasks, even competitive with offline counterparts. The framework design of InfoGrowth allows it to be scaled to an even larger data scale. We hope this new kind of online efficient algorithm will contribute to the creation of high-quality large-scale deep learning datasets and benefit sustainable AI.

\section*{Acknowledgements}
This work is supported by Alibaba Group through Alibaba Innovative Research Program. This work is supported by the National Research Foundation, Singapore under its
AI Singapore Programme (AISG Award No: AISG2-PhD-2021-08-008). Yang You’s research group
is being sponsored by NUS startup grant (Presidential Young Professorship), Singapore MOE Tier-1
grant, ByteDance grant, ARCTIC grant, SMI grant (WBS number: A-8001104-00-00), Alibaba grant,
and Google grant for TPU usage. This work is partially supported by the National Natural Science Foundation of China (62176165), the Stable Support Projects for Shenzhen Higher Education Institutions (20220718110918001), the Natural Science Foundation of Top Talent of SZTU(GDRC202131). This work was supported in part by the National Science and Technology Major Project
(2021ZD0110901). We thank Haonan Wang for valuable discussions and feedback.

\bibliographystyle{splncs04}
\bibliography{main}
\end{document}